CrossMark

# Smooth head tracking for virtual reality applications

Abdenour Amamra[1]

**Abstract** In this work, we propose a new head-tracking solution for human–machine real-time interaction with virtual 3D environments. This solution leverages RGBD data to compute virtual camera pose according to the movements of the user's head. The process starts with the extraction of a set of facial features from the images delivered by the sensor. Such features are matched against their respective counterparts in a reference image for the computation of the current head pose. Afterwards, a prediction approach is used to guess the most likely next head move (final pose). Pythagorean Hodograph interpolation is then adapted to determine the path and local frames taken between the two poses. The result is a smooth head trajectory that serves as an input to set the camera in virtual scenes according to the user's gaze. The resulting motion model has the advantage of being: continuous in time, it adapts to any frame rate of rendering; it is ergonomic, as it frees the user from wearing tracking markers; it is smooth and free from rendering jerks; and it is also torsion and curvature minimising as it produces a path with minimum bending energy.

## 1 Introduction

The availability of high computational power and multi-functional sensors at an affordable cost has encouraged many researchers to look back into the limits of Human–Machine (HM) interaction. Until recently, keyboard and mouse have been the major means of communication with machines. Though useful, the latter are not a natural way of human-to-human communication. For this reason, efforts have been multiplied in order to allow people to interact with machines more naturally via gesture, facial expressions and speech. In the gaming industry, the Microsoft Kinect, Nintendo Wii and PlayStation Move have already shifted the way we play games from manual control to realistic interaction.

In this work, we are interested in free-viewpoint systems. Such systems allow the user to interactively control the position and orientation (both constitute the pose) of the virtual camera looking at a simulated 3D environment. In other words, the user can focus on any desired point in the 3D scene. More importantly, we are trying to make the operation of viewpoint control as ergonomic as possible by tracking the user's focus, instead of asking him to provide it.

To this end, we first need to estimate accurately the pose of the head at a single frame. Such an operation involves facial features extraction and matching for the computation of the 6 DOF pose. The subject features should be discriminative and independent from any facial deformations. In our case, we leverage both colour and depth data delivered by an RGBD sensor working at 30 frames per second (FPS) for a more robust detection and tracking. The result of this stage is a single accurate pose estimation.

Generally, 3D applications require image data to be projected on screens at 60 FPS, which is neatly higher that the native frame rate of the camera (30 FPS maximum). In addition, the operations of feature extraction and matching achieved in the previous stage are time costly. Hence, the algorithm would not be able to deliver a pose for every

✉ Abdenour Amamra
amamra.abdenour@gmail.com

[1] Computer Science Lab., Ecole Militaire Polytechnique, BP17, Bordj El-Bahri, Algiers, Algeria

Springer

captured frame. On the other hand, the user's head moves smoothly and continuously as opposed to the rendered view which follows the refresh rate of the virtual camera. For these reasons, we will adapt a motion reconstruction scheme that interpolates the path between the current and the next pose of the user's head. The former (current pose) is assumed to be known, whereas the latter is completely unknown. The reconstructed trajectory should be smooth, continuous, rotation minimising and truly representative of the real motion.

The remainder of the paper is organised as follows: in the next section, the related works on HM interaction are presented along with the different head-tracking approaches. In the following section, HM interaction with 3D scenes based on head motion is formulated. In the next section, we summarise our tracking approach that would allow us to acquire the displacement of the user's head. The result of this section is an accurate estimation of the current and the next poses. Afterwards, the motion between the two poses is reconstructed with a continuous model. In order to evaluate our contribution, we test it on three publicly available data sets then we provide a detailed discussion of its pros and cons. Lastly, the paper is concluded and potential future works are planned.

## 2 Related works

3D interaction techniques typically involve an operation of selection of a given entity in the 3D scene; navigation, or camera displacement in the simulated world; and manipulation or control, where the virtual objects undergo an action performed by the user [1]. In this work, we are interested in the operation of navigation, where the motion of the user's head is interpreted as a series of transformations applied to the virtual camera. For this reason, the resulting navigation is completely dependent on the quality of head pose estimation.

A plethora of methods regarding head-based HM interaction has been already contributed. One of the major classes of head pose estimation methods is *feature-based* [2]. In other words, just some specific facial features such as the nose or the eyes are considered. These features should be visible across different frames. Vatahska et al. [2] detect the eyes and nose tip with AdaBoost [3] classifiers. The result is fed into a neural network to estimate head orientation. Similarly, Whitehill et al. [4] presented an algorithm that relies on the detection of nose tip and both eyes. Hence, the recognisable poses are limited to the ones where both eyes can be observable. Morency et al. [5] proposed a probabilistic framework called Generalised Adaptive View-based Appearance Model.

The previous works rely on 2D images; hence, they are sensitive to illumination changes and the lack of texture on some parts of the face such as the cheeks and the forehead. The release of the recent depth sensors such as Asus Xition and the Microsoft Kinect has inspired researchers to leverage the extra depth information in the process of head pose estimation. Seemann et al. [6] presented a neural network that fuses skin colour histogram and depth information. Breitenstein et al. [7] proposed another nose-based head-tracking solution capable of handling pose change over time. Cai et al. [8] aligned a 3D face model on an RGBD sequence of frames in order to track the features over time. Fanelli et al. [9] used random decision forests to estimate head pose in real time. Likewise, Wang et al. [10] used point signatures and Gabor filter [11] to detect facial key points in depth images. Their approach requires all the facial features to be visible, thus restricting the freedom of head motion while being sensitive to occlusions. Zhao et al. [12] proposed a 3D Statistical Facial Feature Model that models both the global variations in the morphology of the face and the local structures around the features.

All the above cited methods either compute the pose of the head in a given frame or track its changes across different frames. The utilisation of the resulting transformations in 3D animation applications leads to a jerky unpleasant rending. This problem is due to the discontinuities caused by the low frame rate characterising rudimentary cameras (30 FPS for most consumer cameras). Hence, our motivation to contribute a novel solution emerged that animates the 3D scene smoothly with a continuous motion model generated upon the interpolation of the frame-to-frame head poses.

## 3 Problem statement

Our purpose is the estimation of a smooth trajectory of the user's head. To this end, we utilise a sequence of depth images for the user. These images are then processed and the 6 degrees-of-freedom (DOF) pose is estimated.

Generally, the frame rate of the camera, ranges from 15 to 30 FPS; on the other hand, the graphic rendering engine works at a remarkably higher frame rate easily exceeding 60 FPS. Such a large difference in frame rate yields jerk that in turn compromises the graphic experience of the user. In addition, the irregularity of rendering frame rate may cause another synchronisation issue due to incoherence between sound and image streams.

In order to remedy these problems, we use an tracking algorithm that allows one to take a single snapshot for the user's face, estimate the corresponding head pose, and predict the next move in order to reconstruct the missing motion between the captured pose and the predicted one with a parametric model. The resulting motion is characterised by smoothness, due to its rotation minimising property, regularity and stable sampling rate. The user would, therefore, live a realistic immersion as he sees the scene being updated simultaneously with his head movements.

## 4 Head motion estimation

The representation of face has been studied extensively in the literature [13]. In the present work, we are interested in feature-based pose estimation. After being detected, feature points of the current view are matched against their counterparts in the reference one. Then, we proceed through the estimation of the underlying transformation that aligns the current head pose on a reference one at a known pose using the method presented in [14]. We assume that we have two clouds of 3D points representing the *current* and the *reference* sets of facial landmarks $Q = \{q_1, \ldots, q_n\}$, $P = \{p_1, \ldots, p_n\}$, respectively. Each of the elements $p_i, q_i$ has three coordinates $p_i = (x_p, y_p, z_p)_i$ and $q_i = (x_q, y_q, z_q)_i$. We also assume that the $k$th point $q_k$ in the reference feature set had been already matched with the $k$th point in the current feature set $p_k$.

Pose interpolation procedure requires an initial and a final pose estimate to deliver a typical motion model. For instance, the initial state can be assumed equal to the current pose $x_k$ and the final one $x_{k+1}$ is predicted with a Kalman filter scheme [15]. The path between $x_k$ and $x_{k+1}$ is then reconstructed with the following interpolation method.

## 5 Motion reconstruction

At this level, we assume that we have already determined the initial and the final (predicted) pose of the head $p_i = (x_i, y_i, z_i, \alpha_i, \beta_i, \gamma_i)$ and $p_f = (x_f, y_f, z_f, \alpha_f, \beta_f, \gamma_f)$, respectively. The objective now is the reconstruction of the motion undergone by the head between the two poses. Given a parameter $t$, the trajectory linking these poses is assumed to be:

$$r(t) = (x(t), y(t), z(t)); \quad 0 \leq t \leq 1 \tag{1}$$

$$p_i = r(0) = (x(0), y(0), z(0))$$

$$p_f = r(1) = (x(1), y(1), z(1)) \tag{2}$$

### 5.1 Pythagorean hodographs

Mathematically, the determination of the position on the curve $r(t)$ at time $t_1$ requires the evaluation of the integral:

$$r(t_1) = \int_0^{t_1} r'(t)\mathrm{d}t \tag{3}$$

Which means:

$$r(t_1) = \int_0^{t_1} \sqrt{x'(t)^2 + y'(t)^2 + z'(t)^2}\mathrm{d}t \tag{4}$$

$x'(t), y'(t), z'(t)$ are the derivatives or *hodographs* of the three components $x(t), y(t), z(t)$. As has been claimed in [16], there is no closed form solution for integral (4); Instead, numerical approximations are usually employed. Nevertheless, the elimination of the square root from expression (4) would lead to an exact evaluation. Suppose we have $\sigma(t)$ such that:

$$\sigma(t)^2 = x'(t)^2 + y'(t)^2 + z'(t)^2 \tag{5}$$

$$\text{Then}: \quad r(t_1) = \int_0^{t_1} |\sigma(t)| \, \mathrm{d}t \tag{6}$$

Pythagorean hodographs (PH) [16] can provide a closed form solution to the problem of position determination. PH is the first derivative of a parametric polynomial $r(t)$ that satisfies the Pythagorean condition; in other words, with polynomials $u(t), v(t), p(t), q(t), r' = (x'(t), y'(t), z'(t))$ is assumed to be of the form:

$$\begin{cases} x'(t) = u(t)^2 + v(t)^2 - p(t)^2 - q(t)^2 \\ y'(t) = 2(u(t)q(t) - v(t)p(t)) \\ z'(t) = 2(v(t)q(t) - u(t)p(t)) \end{cases} \tag{7}$$

Hence:

$$\sigma(t) = u(t)^2 + v(t)^2 + p(t)^2 + q(t)^2 \tag{8}$$

The resulting curve in Bezier form becomes:

$$r(t) = \sum_{k=0}^{5} p_k \binom{5}{k} (1-t)^{5-k} t^k \tag{9}$$

With $P_k$ being Bezier control points, $r(t)$ is defined by:

$$r(t) = \begin{bmatrix} 1 \\ t \\ t^2 \\ t^3 \\ t^4 \\ t^5 \end{bmatrix}^T \begin{bmatrix} 1 & 0 & 0 & 0 & 0 & 0 \\ -5 & 5 & 0 & 0 & 0 & 0 \\ 10 & -20 & 10 & 0 & 0 & 0 \\ -10 & 30 & -30 & 10 & 0 & 0 \\ 5 & -20 & 30 & -20 & 5 & 0 \\ -1 & 5 & -10 & 10 & -5 & 1 \end{bmatrix} \begin{bmatrix} P_0 \\ P_1 \\ P_2 \\ P_3 \\ P_4 \\ P_5 \end{bmatrix} \tag{10}$$

The commonly used non-trivial polynomials in literature giving Pythagorean hodographs are cubics and quintics, i.e. of degree three and five, respectively. Cubics are simple to compute but less flexible to represent the motion of a human's head; conversely, quintics are more flexible to represent a realistic motion, but more demanding to compute. Hence, we opted for quintic polynomials in order to interpolate the motion correctly.

The determination of the path requires the computation of Bezier control points. To this end, we use quaternions for a more comprehensive and compact formulation. A spatial quintic PH considers a quadratic polynomial of the form:

$$A(t) = A_0(1-t)^2 + A_1 2(1-t)t + A_2 t^2 \qquad (11)$$

With quaternion coefficients:

$$A_m = u_m + v_m \vec{\imath} + p_m \vec{\jmath} + q_m \vec{k}; \quad m = 0, 1, 2 \qquad (12)$$

$\vec{\imath}$, $\vec{\jmath}$ and $\vec{k}$ are unit vectors in the three directions, $\vec{ox}$, $\vec{oy}$ and $\vec{oz}$, respectively. From (7), the spatial Pythagorean hodograph becomes:

$$
\begin{aligned}
r'(t) =\ & \left(u(t)^2 + v(t)^2 - p(t)^2 - q(t)^2\right)\vec{\imath} \\
& + 2\left(u(t)\,q(t) - v(t)\,p(t)\right)\vec{\jmath} \\
& + 2\left(v(t)\,q(t) - u(t)\,p(t)\right)\vec{k}
\end{aligned} \qquad (13)
$$

Equation (13) can be expressed more simply in quaternion form with:

$$r'(t) = A(t)\, i\, A^*(t) \qquad (14)$$

Such that $A^*$ is the conjugate of $A$.

The coefficients in (11) should first be computed in order to determine control points. The latter (control points) would allow us to find and shape a unique curve (path), whose extremities must verify the constraints at extremal points.

The computation of $A_0$, $A_1$ and $A_2$ can be achieved by solving a Hermit interpolation problem. Bezier control points will then be:

$$p_1 = p_0 + \frac{1}{5} A_0 i A_0^* \qquad (15)$$

$$p_2 = p_1 + \frac{1}{10}\left(A_0 i A_1^* + A_1 i A_0^*\right) \qquad (16)$$

$$p_3 = p_2 + \frac{1}{30}\left(A_0 i A_2^* + 4 A_1 i A_1^* + A_2 i A_0^*\right) \qquad (17)$$

$$p_4 = p_3 + \frac{1}{10}\left(A_1 i A_2^* + A_2 i A_1^*\right) \qquad (18)$$

$$p_5 = p_4 + \frac{1}{5} A_2 i A_2^* \qquad (19)$$

By setting $p_0$ to $p_i$ and $p_5$ to $p_f$, the coefficients $A_0$, $A_1$ and $A_2$ are given as follows:

$$
\begin{aligned}
A_0 = \sqrt{0.5\left(1 + \tilde{\alpha}_i\right)|d_i|}\ \bigg( & -\sin(\varphi_0) + \cos(\varphi_0)\,\vec{\imath} \\
& + \frac{\tilde{\beta}_i \cos(\varphi_0) - \tilde{\gamma}_i \sin(\varphi_0)}{1 + \tilde{\alpha}_i}\,\vec{k} \\
& + \frac{\tilde{\beta}_i \cos(\varphi_0) + \tilde{\gamma}_i \sin(\varphi_0)}{1 + \tilde{\alpha}_i}\,\vec{\jmath} \bigg)
\end{aligned} \qquad (20)
$$

$$
\begin{aligned}
A_2 = \sqrt{0.5\left(1 + \tilde{\alpha}_f\right)|d_f|}\ \bigg( & -\sin(\varphi_2) + \cos(\varphi_2)\,\vec{\imath} \\
& + \frac{\tilde{\beta}_f \cos(\varphi_2) + \tilde{\gamma}_f \sin(\varphi_2)}{1 + \tilde{\alpha}_f}\,\vec{\jmath} \\
& + \frac{\tilde{\beta}_f \cos(\varphi_2) - \tilde{\gamma}_f \sin(\varphi_2)}{1 + \tilde{\alpha}_f}\,\vec{k} \bigg)
\end{aligned} \qquad (21)
$$

$$
\begin{aligned}
A_1 = -\frac{3}{4}\left(A_0 + A_2\right) + \frac{1}{4}\sqrt{0.5\left(1 + \tilde{\alpha}\right)|g|}\ \bigg( & -\sin(\varphi_1) \\
& + \cos(\varphi_1)\,\vec{\imath} + \frac{\tilde{\beta}\cos(\varphi_1) + \tilde{\gamma}\sin(\varphi_1)}{1 + \tilde{\alpha}}\,\vec{\jmath} \\
& + \frac{\tilde{\beta}\cos(\varphi_1) - \tilde{\gamma}\sin(\varphi_1)}{1 + \tilde{\alpha}}\,\vec{k} \bigg)
\end{aligned} \qquad (22)
$$

$$
\begin{aligned}
\vec{g} = 120 * \left(p_f - p_i\right) - 15\left(d_f + d_i\right) + 5\big(A_0 i\, A_2^* \\
+ A_2 i\, A_0^*\big)
\end{aligned} \qquad (23)
$$

where, $d_i = \left(\alpha_i, \beta_i, \gamma_i\right)$ and $d_f = \left(\alpha_f, \beta_f, \gamma_f\right)$ and $\left(\tilde{\alpha}_i, \tilde{\beta}_i, \tilde{\gamma}_i\right)$ and $\left(\tilde{\alpha}_f, \tilde{\beta}_f, \tilde{\gamma}_f\right)$ are their respective direction cosines. $\tilde{\alpha}$, $\tilde{\beta}$, $\tilde{\gamma}$ are the direction cosines of the vector $\vec{g}$. The two angles $\varphi_1$, $\varphi_2$ are free parameter that can be all a solution for the interpolation problem. The difference between the solutions corresponding to different values of the two angles lies in the value of curvature that significantly varies from one combination to another. As has been claimed in [16], when we set $\varphi_1 = \varphi_2 = -\frac{\pi}{2}$ we obtain a rotation minimising frame (RMF).

## 5.2 Curvature and Torsion constraints

By its very nature, the human head has a limited interval of freedom of rotation around the three principle vectors constituting its local frame (tangent, normal, binormal). The interpolated curve must have a curvature and torsion that fall below certain tolerable threshold. The latter is defined by the biomechanical properties of the movement of the human head.

The resulting trajectory has a tangential continuity, but it should have a curvature ($\kappa(t)$) small enough to allow the head movements to be feasible, in other words:

$$|\kappa(t)| < \kappa_{\max} \qquad (24)$$

$\kappa_{\max}$ is the maximum tolerable curvature that can be undergone by the moving head. Setting the two free variables $\varphi_1$, $\varphi_2$ to $\varphi_1 = \varphi_2 = -\frac{\pi}{2}$ results in a minimum curvature as has been argued previously.

On the other hand, the torsion ($\tau(t)$) of the path followed by the head should also be bounded with a maximum value that ensures a natural behaviour along the curve.

$$|\tau(t)| < \tau_{max} \qquad (25)$$

Torsion governs the rotation of the frame along the path; its computation depends only on local properties of the interpolated curve ($r(t)$). In order to minimise the rotation of the local frame, we readjust the shift between the Frenet-Serret (FR) [17] frame and the one with minimum rotation (PH) by rotating the Normal and Binormal vectors with $\theta(t)$. Such an angle will compensate for the useless point-to-point rotations as follows:

$$\theta(t) = -\int_0^t \tau(s)\,|r'(s)|\,ds \qquad (26)$$

The application of frame orientation readjustment leads to the minimisation of the total rotation over the reconstructed path. The corrective effect of our approach can be seen on flection points where PH frames change their orientation slightly and smoothly, whereas FR frames suddenly flip with $\frac{\pi}{2}$. When applied on 3D rendering, this abrupt change causes the whole scene to turn with half a circle, though the user's head has just passed through a flection point.

# 6 Result and discussions

In order to assess our contribution, we carried out a series of experiments on three datasets for head, object and camera tracking. For each, we compared position and orientation root-mean-square error (RMSE) (measure of accuracy), curvature and torsion (measures of comfort) delivered by our algorithm against those of the state-of-the-art solutions analysed in the benchmarks.

## 6.1 Datasets description

### 6.1.1 Head-tracking dataset (Head)

We used a head pose estimation dataset [9] with a variety of RGBD images for human faces along with their ground-truth 6 DOF pose data. The dataset contains over 15,000 frames for 20 people; each frame includes a depth map and its corresponding RGB image ($640 \times 480$ pixels) along with ground-truth annotation. The data have been taken for different persons sitting in front of the Kinect; the subjects were asked to freely turn their head around, trying to span all possible pitch/yaw angles they could reach. The recorded sequences were annotated with the position of the head and its three orientation angles. The particularity of this dataset is that the authors do not capture an image at every frame, they rather capture one at random time steps. This property affects significantly the reconstruction process due to the lack of data for relatively long distances; hence its suitability to assess our contribution objectively.

### 6.1.2 Object-tracking dataset (Object)

The authors of this dataset [18] recorded 100 video clips using a Kinect sensor at resolution ($640 \times 480$ pixels). The images were taken indoors due to the poor performance of the camera outdoor. The depth of objects ranges from 0.5 to 10 metres. The authors manually annotated target objects location by drawing a minimum bounding box surrounding them on each frame. When the target moves, this box is readjusted automatically to fit its visible region. When the target becomes completely occluded, the bounding box is removed. The targets can be humans, animals or rigid objects. Here we restrict ourselves to rigid objects.

### 6.1.3 Camera-tracking dataset (Camera)

Camera-tracking dataset [19] contains colour and depth images captured by a Kinect along the ground-truth trajectory of the sensor. The data was recorded at full frame rate (30 FPS) and sensor resolution ($640 \times 480$ pixels). The ground-truth trajectory was obtained from a high-accuracy motion capture system with eight high-speed tracking cameras. The recorded trajectories were taken by a handheld Kinect in a typical office environment ($6 \times 6\,m^2$) as well as a forward looking camera mounted on a wheeled robot driving in a large industrial hall ($10 \times 12\,m^2$). The captured data was basically intended to demonstrate the utilisation of the Kinect in SLAM (Simultaneous Localisation And Mapping) applications.

## 6.2 Performance assessment metrics

We choose a subset of image data from each of the aforementioned datasets, and then we run our algorithm (PH) against three state-of-the-art algorithms tested on the benchmarks ($H_1$ [9], $H_2$ [20], $H_3$ [21], for Head, Object and Camera, respectively). The validation is threefold: first, we compute position/rotation RMSEs that separate the results from the ground truth. Such a metric allows us to assess the accuracy of the estimated trajectory. Second, we compute the curvature of our estimated trajectories and that of the result delivered by alternative methods. For instance, the curvier the trajectory, the less ergonomic it is. Hence, we are interested in trajectories with the least amount of curvature. Third, we assess the torsion characterising the resulting trajectories. The minimisation of this metric is also desirable as we want to reduce the rotations around the tangent vector when we browse the resulting path. An example of the undesirable effect of high level of torsion in the context of 3D gaming and simulation is motion sickness; the latter results from the significant camera rotations in a short amount of time.

The data of each benchmark was divided into 400 samples of consistent sequences ordered from the least (corresponding to 1) to the most (corresponding to 400) challenging

SIViP

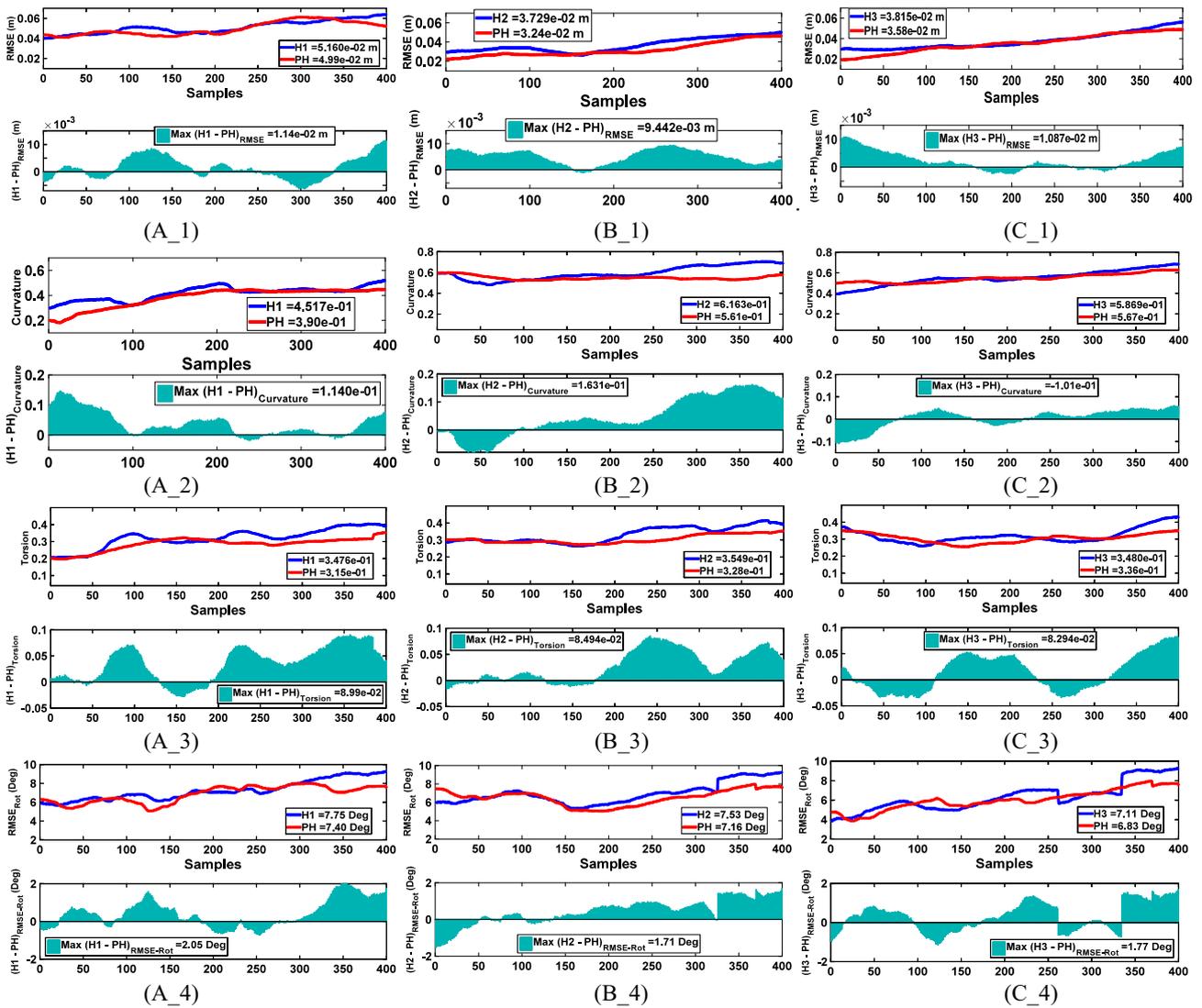

**Fig. 1** *Columns* (one dataset per column): **a** *Head*, **b** *Object*, **c** *Camera*. *Rows* (one metric per row): (_1) Position RMSE in metres, (_2) estimated trajectory average curvature, (_3) estimated trajectory average torsion, (_4) rotation RMSE in degrees. The *red curves* represent the performance of our solution (PH); whereas, the *blue* ones represent the corresponding results obtained from state-of-the-art algorithms, i.e. $H_1$ for *Head*, column (**a**); $H_2$ for *Object*, column (**b**) and $H_3$ for *Camera*, column (**c**) (colour figure online)

conditions. For instance, each series of images in the same sequence belongs to a single scene. For each sequence, we compute PH and $H_{i,1 \leq i \leq 3}$ RMSE, curvature and torsion; then we plot their respective graphs. In parallel, we visualise the contrast of performance between PH and $H_{i,1 \leq i \leq 3}$.

### 6.3 Results

As can be seen in Fig. 1, every dataset corresponds to one of the three columns, i.e. Fig. 1_a corresponds to Head, Fig. 1_b corresponds to Object and Fig. 1_c corresponds to Camera, then every line represents a single metric, i.e. position RMSE (line _1), curvature (line _2), torsion (line _3) and rotation RMSE (line _4).

### 6.3.1 Head

As can be seen in Fig. 1a_1, head position RMSE ranges from 0.04 to 0.06 m. On average, our algorithm's RMSE (PH in red) was ~0.05 m, as opposed to $H_1$ (in blue), which gave 0.052 m. This amounts to an advantage, RMSE($H_1$-PH), of 0.002 m in favour of PH. The largest gap, Max(RMSE($H_1$-PH)), was 0.011 m. Curvature, on the other hand (see Fig. 1a_2), ranges from 0.2 to 0.43 for both PH and $H_1$ over the set of samples. On average, PH curvature was 0.39, whereas $H_1$'s was 0.45. The difference, curvature ($H_1$-PH), was 0.06, and the largest gap was 0.11. Likewise, torsion (see Fig. 1a_3) ranges from 0.2 to 0.4. Average PH torsion was 0.31, whereas $H_1$'s was 0.35. The difference, torsion($H_1$-

PH), was 0.04, and the largest gap was ∼0.09. Finally, head rotation RMSE (see Fig. 1a_4) fluctuates between $4°$ and $10°$ with an average of $7.40°$ for PH and $7.75°$ for $H_1$. The largest gap between the two algorithms was $2.05°$.

### 6.3.2 Object

In Fig. 1b, we illustrate tracking results when the subject is a rigid object. Object position RMSE ranges from 0.02 to 0.05 m. On average, PH's RMSE (red in Fig. 1b_1) was 0.032 m and $H_2$'s (in blue) was 0.037 m. The difference between the two, i.e. RMSE($H_2$- PH) was 0.005 m, resulting in an almost equivalent performance for both algorithms. The Max(RMSE($H_2$-PH)) was 0.01 m. Curvature (Fig. 1b_2), varies slightly around 0.6. On average, PH curvature was 0.56, whereas $H_2$'s was 0.67. The difference, curvature($H_2$-PH), was 0.01, and the largest gap was 0.163. Torsion also varies slightly around 0.3 (see Fig. 1b_3). Average PH's torsion was 0.33, whereas $H_2$'s was 0.35; this yields an average difference, i.e. torsion($H_2$-PH), of 0.02, and a maximum gap of 0.085. As with Head-tracking benchmarks, object rotation RMSE (see Fig. 1b_4) fluctuates between $4°$ and $10°$ with an average of $7.16°$ for PH and $7.53°$ for $H_2$. The largest recorded gap between PH and $H_2$ was $1.71°$.

### 6.3.3 Camera

Figure 1c_1 illustrates position RMSE when the tracked entity is the camera itself. Overall, its (RMSE) ranges from 0.02 to 0.06 m. On average, PH (red) gave an RMSE of 0.036 m, whereas $H_3$ (in blue) gave 0.038 m. The difference, RMSE($H_3$-PH), was 0.002 m in favour of PH. The largest gap, Max(RMSE($H_3$-PH)), was 0.01 m. Curvature (see Fig. 1c_2), ranges from 0.4 to 0.6 for $H_3$ and it stays around 0.5 for PH. On average, PH curvature was 0.56, whereas $H_3$'s was 0.58, which leads to a difference, curvature($H_3$-PH), of 0.02, and a largest gap of 0.1 in favour of $H_3$. Torsion (see Fig. 1c_3) varied between 0.25 and 0.4. Average PH torsion was 0.34; whereas $H_3$'s was 0.35. The average difference, torsion($H_3$-PH), was 0.01, and the largest gap was 0.083. Camera rotation RMSE for Camera dataset (see Fig. 1c_4) also fluctuates between $4°$ and $10°$ with an average of $6.83°$ for PH and $7.11°$ for $H_3$, leading to an average difference of $0.28°$. The maximum gap between PH and $H_3$ was $1.77°$.

### 6.3.4 Processing time

We also computed the average processing time for different levels of details (number of reconstructed points in a segment, the higher this number the most precise the reconstruction). Computation time scales linearly with the size of the segment. Such a advantage qualifies our solution for use in real-time applications as well as high-resolution tracking where it would consequently lead to a smoother and more decent rendering.

## 7 Conclusion and future works

We proposed a novel smooth head-tracking solution for human–machine real-time interaction with virtual environments. We extracted a set of facial features from the RGBD data delivered by a depth sensor. These features are matched against their respective counterparts in a reference image, and then the current head pose is computed (initial pose). We also used a prediction approach in order to determine the next head move (final pose). More importantly, we proposed a Pythagorean hodograph-based strategy for head motion reconstruction between the initial and the final poses. This motion can be used directly for the animation of a 3D scene in front of the tracked user's face.

Test results, carried out on Head, Object and Camera benchmarks, show position and rotation accuracy of the proposed method. More importantly, it minimises the total curvature of the reconstructed trajectory as well as the rotation of the frame around the tangent vector. These characteristics lead to a high level of comfort of navigation in 3D environment. In addition, our head-tracking solution is ergonomic as it does not require the user to actively provide commands that will be interpreted as 3D controls in the virtual scene. Moreover, execution time demonstrates its adaptability to the desired rendering frame rate.

As a future work, we intend to exploit our findings in the development of a complete Human–Machine interaction system. Furthermore, we would investigate the Pythagorean hodographs more deeply for prospective resolution of more computer vision problems.